\pdfoutput=1 
\documentclass{cta-author}
\usepackage[utf8]{inputenc}

\usepackage{amsmath,amssymb,amsfonts}
\usepackage{graphicx}
\usepackage{textcomp}
\usepackage{caption}
\usepackage{diagbox}
\usepackage{colortbl}
\usepackage{multirow}
\usepackage{multicol}
\usepackage{booktabs}
\usepackage{tabulary}
\usepackage{caption}

\usepackage{algorithmicx}
\usepackage{algorithm}
\usepackage{algpseudocode}

\definecolor{mygray}{gray}{.9}
\definecolor{mypink}{rgb}{.99,.91,.95}
\definecolor{mygreen}{rgb}{.70,.93,.70}
\definecolor{mycyan}{cmyk}{.3,0,0,0}

{}
{}
{}
\begin{document}

\supertitle{Submission Template for IET Research Journal Papers}

\title{\LARGE SASO: Joint 3D Semantic-Instance Segmentation via Multi-scale Semantic Association and Salient Point Clustering Optimization}

\author{\au{Jingang Tan$^{1,2}$}, \au{Lili Chen$^{1,2*}$}, \au{Kangru Wang$^{1,2}$},\au{Jingquan Peng$^{1,2}$},\au{Jiamao Li$^{1,2}$},\au{Xiaolin Zhang$^{1,2,3}$}}

\address{
\add{1}{Bionic Vision System Laboratory, State Key Laboratory of Transducer Technology, Shanghai Institute of Microsystem and Information Technology, Chinese Academy of Sciences, Shanghai, China.}
\add{2}{University of Chinese Academy of Sciences, Beijing, China.}
\add{3}{School of Information Science and Technology, ShanghaiTech University, Shanghai, China.}
\email{lilichen@mail.sim.ac.cn}}

\begin{abstract}
We propose a novel 3D point cloud segmentation framework named \textit{SASO}, which jointly performs semantic and instance segmentation tasks. For semantic segmentation task, inspired by the inherent correlation among objects in spatial context, we propose a Multi-scale Semantic Association (\textit{MSA}) module to explore the constructive effects of the semantic context information. For instance segmentation task, different from previous works that utilize clustering only in inference procedure, we propose a Salient Point Clustering Optimization (\textit{SPCO}) module to introduce a clustering procedure into the training process and impel the network focusing on points that are difficult to be distinguished. In addition, because of the inherent structures of indoor scenes, the imbalance problem of the category distribution is rarely considered but severely limits the performance of 3D scene perception. To address this issue, we introduce an adaptive Water Filling Sampling (\textit{WFS}) algorithm to balance the category distribution of training data.
Extensive experiments demonstrate that our method outperforms the state-of-the-art methods on benchmark datasets in both semantic segmentation and instance segmentation tasks.
\end{abstract}

\maketitle
\section{Introduction}\label{sec1}
Scene perception plays a decisive role in many applications, such as autonomous driving, robot navigation and augmented reality. With the growth of computer technology and artificial intelligence in recent years, scene perception ability of intelligent devices has received increasing attention from both academia and industry, especially for the 3D scenes which can represent the real environment intuitively. Semantic segmentation and instance segmentation of 3D scenes are the fundamental and critical portions of 3D scene perception.
Nevertheless, how to model the 3D space into digital shape to accomplish scene segmentation task is an indefinite problem.
Various representations of 3D scenes have been investigated, such as depth maps, voxels, multi-views, meshes and point clouds.
Based on these representations, a series of excellent works have been investigated to operate segmentation task,
such as \cite{wang2019voxsegnet,dai20183dmv,qi2017pointnet,qi2017pointnet++,engelmann2017exploring,graham20183d,shen2018mining,huang2018recurrent,Xiaoqing20183D,yi2019gspn,yang2019learning,lahoud20193d,liu2019masc,wang2018sgpn,wang2019associatively,pham2019jsis3d,liang20193d,hou20193d}. Among these representations, point clouds are the most compact and natural to the geometric distributions of real 3D scenes, which have been applied extensively in recent researches. 
In terms of semantic and instance segmentation tasks in 3D point clouds, based on the great success achieved in recent years \cite{Landrieu2018Large,wang2019voxsegnet,graham20183d,wang2019graph,dai20183dmv,engelmann2017exploring,Xiaoqing20183D,wang2018sgpn,yi2019gspn,yang2019learning,lahoud20193d,liu2019masc,elich20193d} for each single task, joint learning methods for both tasks \cite{wang2018sgpn,pham2019jsis3d, wang2019associatively} have opened up a new effective way to explore the 3D scene segmentation, which improved the performance and promoted further development.
Compared with the method \cite{wang2018sgpn} exploiting similarity matrix, \cite{pham2019jsis3d, wang2019associatively} utilized clustering algorithm to generate instance segmentation result, which was proved to be more effective and flexible. Nevertheless, whether the convergence direction of the training process is consistent with the orientation of clustering algorithm was rarely considered. Additionally, the marginal points are usually harder to be distinguised than the central points, and in multiple objects case the internal points are easier to be distinguished than the boundary points across objects, as shown in Figure \ref{fig3}. To address this problem, we propose a Salient Point Clustering Optimization (\textit{SPCO}) module to introduce clustering into the training process and saliently focus on the points that are harder to be distinguished in the clustering process. As for semantic segmentation, the spatial distribution of the semantic information has a strong association, which can be further exploited. For example, when a point comes from table, it is highly possible that there will be some neighbor points belonging to chair other than from ceiling. The most common approach to explore the semantic associations is the Conditional Random Fields (CRF) algorithm \cite{lafferty2001conditional}, which utilizes normalization based on statistical global probability and has been proved to be effective in segmentation tasks. However, CRF is complex and consumes plenty of resources, how to sufficiently exploit the semantic associations more efficiently is an indefinite problem.
Consequently, we propose a Multi-scale Semantic Association (\textit{MSA}) module to fine tune the semantic segmentation results, which is based on the multiple scale semantic association maps generated by statistical analysis. In addition, because of the inherent structures of indoor scenes, the imbalance problem of the category distribution badly limits the performance of 3D scene perception. For example, wall and floor certainly exist in every room while other categories may not, such as sofa, sink, bookshelf, \emph{etc.}. This leads to the numbers of points from wall and floor are much more than the one from other categories. The imbalance problem is rarely considered in previous wokrs. Thus, we present an adaptive Water Filling Sampling (\textit{WFS}) algorithm to address this problem by changing the sampling probabilities of each category adaptively. \\
To summarize, our contributions are the following: 
\begin{itemize}
\item We propose a Salient Point Clustering Optimization (\textit{SPCO}) module to introduce clustering into the training process and saliently focus on the points that are harder to be distinguished in instance segmentation.
\item We propose a Multi-scale Semantic Association (\textit{MSA}) module based on statistical knowledge to explore the potential spatial association of the semantic information in point clouds.
\item We propose an adaptive Water Filling Sampling (\textit{WFS}) algorithm to balance category distribution in the point clouds, which is rarely considered but critical in 3D scene perception. 
\item Extensive experiments demonstrate that our \textit{SASO} outperforms the state-of-the-art related methods on benchmark datasets in both semantic and instance segmentation criteria.
\end{itemize}
\section{Related Works}\label{sec2}
This section reviews recent deep learning-based techniques applied to 3D point clouds. In recent years, a series of deep learning architectures have been proposed to perform the encoding and decoding for 3D point clouds or its derived representations, which are widely utilized in many 3D vision tasks such as semantic and instance segmentation, object part segmentation and object detection. We divide these methods into four categories based on the data representations. Further more, we will introduce recent 3D semantic and instance segmentation research progress based on above techniques.
\subsection{Volumetric Methods}
Due to 3D point clouds are irregular, the most simple but naive method is to voxelize the irregular point clouds to regular 3D grids so that 3D convolutions can be applied \cite{wu20153d,wang2019voxsegnet,graham20183d,wang2017cnn,zhou2018voxelnet,qi2016volumetric,klokov2017escape,riegler2017octnet,lei2019octree,ren2018sbnet,maturana2015voxnet,huang2016point}. Specifically, Wu \emph{et al.} \cite{wu20153d} represented a geometric 3D shape as a probability distribution of binary variables on a 3D voxel grid, using a Convolutional Deep Belief Network. Maturana \emph{et al.} \cite{maturana2015voxnet} proposed an architecture to efficiently deal with large amounts of point cloud data by integrating a volumetric Occupancy Grid representation with a supervised 3D Convolutional Neural Network. Zhou \emph{et al.} \cite{zhou2018voxelnet} removed the manual feature engineering for 3D point clouds and divided point clouds into equally spaced 3D voxels, then transformed a group of points within each voxel into a unified feature representation through a newly introduced voxel feature encoding layer. Wang \emph{et al.} \cite{wang2019voxsegnet} designed a spatial dense extraction module to preserve the spatial resolution during the feature extraction procedure, alleviating the loss of detail caused by sub-sampling operations such as max-pooling.
Although volumetric data representation is the most common and simplest form, there is an obvious drawback that cubic complexity of 3D convolutions leads to a dramatic increase in the memory consumption and computing resources. To tackle this issue, \cite{wang2017cnn,riegler2017octnet} proposed octree representation to improve efficiency of network and reduce computing resources. In addition, \cite{graham20183d,ren2018sbnet} proposed sparse convolutional operations to process spatially-sparse 3D point clouds and achieved impressive results. Although these methods try to alleviate the efficiency problem, they are much more complex than volumetric CNNs and can not fundamentally solve the memory consumption problem.
\begin{figure}[t]
    \begin{center}
     \includegraphics[width=8.5cm]{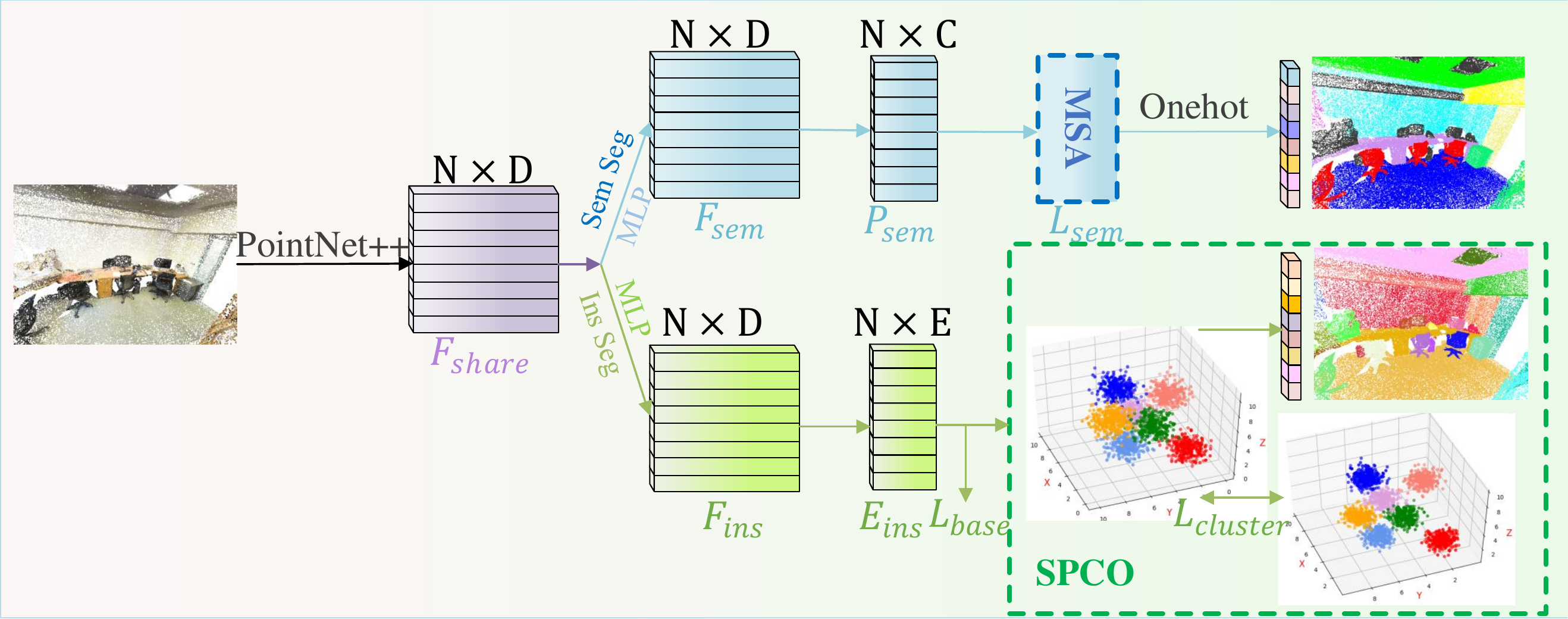}
    \end{center}
    \caption{An illustration of our joint learning framework. The input 3D point clouds are first encoded to $F_{share}$ by PointNet++ \cite{qi2017pointnet++}, then the common feature will be decoded separately by semantic and instance segmentation branches. In semantic segmentation branch (blue), a \textit{MSA} module based on statistics knowledge is proposed to explore the semantic association, we will expound it in Sec \ref{subsec:msa}. For the instance segmentation (green), we proposed \textit{SPCO} module to introduce clustering into the training process and focus on hard-distinguished points, which will be explained in  Sec \ref{subsec:spco}.}
    \label{fig1}
\end{figure}
\subsection{Multi-view Methods}
Another common method for 3D point clouds are a multi-view representation. In recent years, Convolutional Neural Networks have been proved successful in a wide range of 2D visual tasks. To sufficiently take advantage of the strong extraction capability of classical CNNs, 3D point clouds are first projected into multiple pre-defined views, which are then processed by well-designed image-based CNNs to extract features, such as \cite{su2015multi,shi2015deeppano,roveri2018network,you2018pvnet,dai20183dmv,guerry2017snapnet,qi2016volumetric}. Specifically, Guerry \emph{et al.} \cite{guerry2017snapnet} used 3D-coherent synthesis of scene observations and mixed them in a multi-view framework for 3D labeling. Su \emph{et al.} \cite{su2015multi} presented a novel CNN architecture that combines information from multiple views of a 3D shape into a single and compact shape descriptor offering even better recognition performance. Dai \emph{et al.} \cite{dai20183dmv} encoded the sparse 3D point clouds with a compact multi-view representation, including bird’s eye view and front view as well as RGB image to perform high-accuracy 3D object detection. You \emph{et al.} \cite{you2018pvnet} proposed PVNet to integrate both the point cloud and the multi-view data towards joint 3D shape recognition.
Although the multi-view representation of point cloud data is reasonable, the project process from 3D to 2D will loss the full utilization of  3D geometric information.
\subsection{Graph Convolution Methods}
Graph structure is a native representation of irregular data, such as 3D point clouds, which offers a compact yet rich representation of contextual relationships between points of different object parts \cite{bruna2013spectral,wang2018local,te2018rgcnn,simonovsky2017dynamic,wang2019graph,Landrieu2018Large}. Specifically, Bruna \emph{et al.} \cite{bruna2013spectral} proposed two constructions based on a hierarchical clustering of the domain and the spectrum of the graph Laplacian, to prove that for low-dimensional graphs, it is possible to learn convolutional layers with a number of parameters independent of the input size, resulting in efficient deep architectures. Wang \emph{et al.} \cite{wang2018local} operated spectral graph convolution on a local graph, combined with a novel graph pooling strategy to augment the relative layout of neighboring points as well as their features. Te \emph{et al.} \cite{te2018rgcnn} treated features of points in a point cloud as signals on graph, and defined the convolution over graph by Chebyshev polynomial approximation leveraging on spectral graph theory. They also designed a graph-signal smoothness prior in the loss function to regularize the learning process. Although the graph convolutional methods have achieved significant performance, these methods constructed on Laplacian matrix, is computationally complex for Laplacian eigen-decomposition and has a large quantity of parameters to express the convolutional filters while lacks spatial localization.

\subsection{Point clouds Methods}
Point clouds are an intuitive, memory-efficient 3D representation which is well-suited for representing geometric details. How to apply deep learning techniques in point clouds directly, simply and efficiently is a critical problem. To address this challenge, Qi \emph{et al.} \cite{qi2017pointnet} designed a novel type of neural network PointNet that directly consumes point clouds and well respects the permutation invariance of points in the input. More specifically, they solved the disorder problem of the point clouds through max pooling and maintained the rotation invariance through the spatial transformation network STN. The extracted features of each point are the combination of its own information and the global information. PointNet has been proved efficient in many applications ranging from object classification, part segmentation, object detection to scene semantic parsing. However, PointNet only relies on the max-pooling layer to learn global features and does not consider local relationships. Therefore, a series of works \cite{qi2017pointnet++,huang2018recurrent,wang2019dynamic,li2018pointcnn,engelmann2017exploring} were developed through investigations of the local context and hierarchical learning structures. Typically, Qi \emph{et al.} \cite{qi2017pointnet++} proposed PointNet++ based on their previous work PointNet, which utilizes pointnet as a local feature extraction module to operate hierarchical feature extraction like CNNs, and finally uses upsampling to generate the final high level features.
Li \emph{et al.} \cite{li2018pointcnn} proposed PointCNN which uses MLP to learn a transformation matrix to solve the disorder problem of point cloud, and then utilizes the introduced x-conv module to perform convolution on the transformed features. This method achieved similar performance as PointNet++.
\subsection{3D semantic and instance segmentation}
Recent advances in learning-based techniques have also led to various cutting-edge 3D semantic and instance segmentation approaches \cite{Landrieu2018Large,te2018rgcnn,wang2019graph,wang2019voxsegnet,graham20183d,qi2016volumetric,dai20183dmv,qi2017pointnet,qi2017pointnet++,engelmann2017exploring,shen2018mining,li2018pointcnn,hua2018pointwise,huang2018recurrent,wu2019pointconv}. Volumetric representation has been adapted by \cite{wang2019voxsegnet,graham20183d} to transfer 3D point clouds to regular grids and operate CNNs to extract features.
\cite{Landrieu2018Large,te2018rgcnn,wang2019graph} utilized graph convolutional networks to model the relationships of 3D points which offers a compact yet rich representation of context. \cite{qi2016volumetric,dai20183dmv} transfered 3D point clouds into multiple views to sufficiently take advantage of the strong extraction capability of classical CNNs. \cite{qi2017pointnet,qi2017pointnet++,engelmann2017exploring,shen2018mining} presented more efficient and flexible ways to utilize MLP directly upon point clouds and well respect the permutation invariance of points. \cite{li2018pointcnn,hua2018pointwise,wu2019pointconv} operated segmentation task by designing novel CNNs on point clouds, Huang \emph{et al.} \cite{huang2018recurrent} and Ye \emph{et al.} \cite{Xiaoqing20183D} proposed new approaches by slicing the point clouds and utilizing recurrent neural networks to exploit the inherent contextual features.
3D instance segmentation is a relatively new research area and attracts more and more attention \cite{yang2019learning,lahoud20193d,liu2019masc}. Specifically, Lahoud \emph{et al.} \cite{lahoud20193d} proposed a network based on 3D voxel grids, which treats the instance segmentation task as multi-task learning problem. The network generates abstract feature embeddings for voxels and estimates instances' centers to learn instance information. Yang \emph{et al.} \cite{yang2019learning} introduced a framework which simultaneously generates 3D bounding boxes and predicts the binary masks for the points within each box in one stage.
Recently, Wang \emph{et al.} \cite{wang2018sgpn} have opened up a framework by jointly operating semantic and instance segmentation in 3D point clouds. Inspired by the proposal mechanism in 2D FasterRcnn\cite{ren2015faster}, they proposed similarity matrix indicating the similarity between each pair of points in embedded feature space to predict point grouping proposals, then the network will predict corresponding semantic class for each proposal to generate the final semantic-instance results. Although the similarity matrix is effective and natural to indicate the proposals, it will generate a large and inefficient matrix which suffers from the heavy computation and memory consumes. Some followed proposal methods \cite{yi2019gspn,hou20193d} were proposed to boost the performance of similar framework while still depended on two-stage procedure and the time-consuming non-maximum suppression algorithm. 
More recently, \cite{wang2019associatively,pham2019jsis3d} utilized clustering algorithm to divide points into different objects, which was demonstrated to be more effective and efficient than proposal methods. Nevertheless, they did not consider whether the convergence direction of the training process is coupled with the orientation of clustering algorithm. In addition, different points have various diffculties to be devided into distinct objects, which is rarely considered. In this work, we propose a framework which take this critical problem into consideration and prove that it is significant and effective.
\begin{figure}[t]
  \begin{center}
   \includegraphics[width=8.5cm]{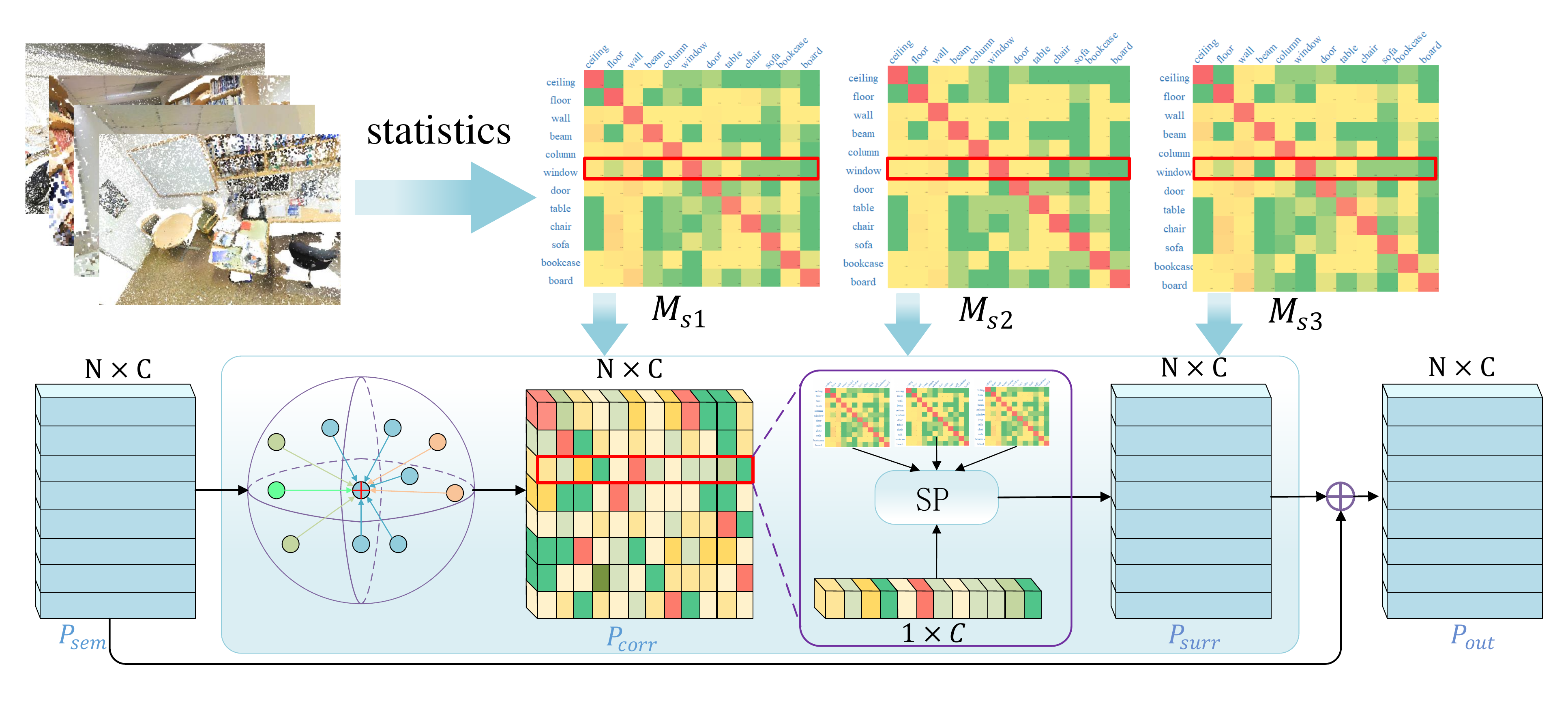}
  \end{center}
  \caption{An illustration of our \textit{MSA} module. First, we create multi-scale semantic association map $M_s$ by statistics with ball query upon all the training 3D scenes. For a point $i$ in the semantic prediction result, we also generate a vector $P_{corr}(i)$ indicating the probabilities of different categories about surrounding points with ball query. Then we calculate the similarity between this vector and each line (category) in the $M_s$ and normalize it as a probability vector, the detail of the calculation $SP$ is formulated in equation \ref{eq8}. The final prediction for each point is the fusion of original predict probability and the fine-tuned probability, as formulated in equation \ref{eq9}.}
  \label{fig2}
\end{figure}
\section{Proposed Method}\label{sec3}
\begin{figure}[t]
  \begin{center}
   \includegraphics[width=8.5cm]{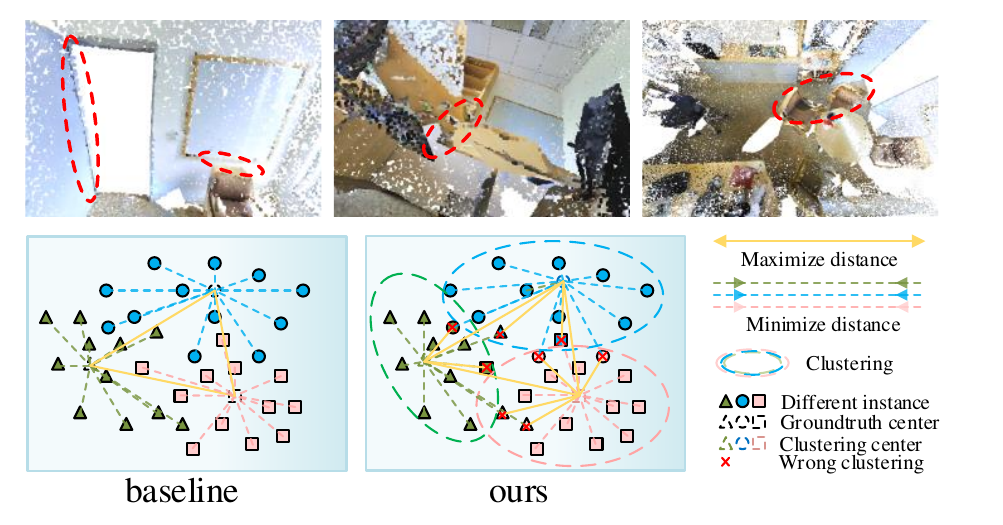}
  \end{center}
  \caption{An illustration of our \textit{SPCO} module. As shown in the first line, the different points of one object have different difficulties to be distinguished, especially for the points of the joints among different objects. In terms of this problem, we introduce clustering into training procedure and saliently focus on the points that are harder to be distinguished in the clustering process.}
  \label{fig3}
\end{figure}

In this section, we first introduce the baseline framework of our network which jointly perform semantic and instance segmentation tasks. Then we give the details of our \textit{MSA} module for semantic segmentation in Sec \ref{subsec:msa}, as depicted in Figure \ref{fig2}. Next, we expound our \textit{SPCO} module for instance segmentation in Sec \ref{subsec:spco}, as shown in Figure \ref{fig3}. The whole framework of our method can be seen in Figure \ref{fig1}. Finally, the adaptive Water Filling Sampling (\textit{WFS}) algorithm is explained in details in Sec \ref{subsec:wfs}.
\subsection{Baseline Framework}
\label{issue}
As depicted in Figure \ref{fig1}, the network without \textit{MSA} and replaced \textit{SPCO} with normal clustering is the baseline framework.
First, point clouds of size $N$ are encoded into a high-dimensional feature matrix $F_{share} \in \mathbb{R}^{N \times D}$ by the encoder PointNet++ \cite{qi2017pointnet++}.
Next, two tasks separately decode $F_{share}$ for their own missions. In the semantic segmentation branch, $F_{share}$ is decoded into the semantic feature matrix $F_{sem} \in \mathbb{R}^{N \times D}$ and then outputs the semantic predictions $P_{sem} \in \mathbb{R}^{N \times C}$, where $C$ is the semantic class number.
The instance segmentation branch decodes $F_{share}$ into the instance feature matrix $F_{ins} \in \mathbb{R}^{N \times D}$, which is utilized to predict the per-point instance embeddings $E_{ins} \in \mathbb{R}^{N \times E}$, where $E$ denotes the length of the output embedding dimensions. These embeddings are used to calculate the distances among the points for instance clustering.
During the training process, the semantic branch is supervised by cross entropy loss while the loss function for instance segmentation, inspired by \cite{wang2019associatively}, is formulate as follows:
\begin{equation}
 \begin{array}{l}
  {\mathcal{L}_{base} = \mathcal{L}_{in} + \mathcal{L}_{out} + \lambda \mathcal{L}_{reg}}
 \end{array}
 \label{eq1}
\end{equation}
where the goal of $\mathcal{L}_{in}$ is to pull the embeddings toward the mean embedding of the points in the instance, while $\mathcal{L}_{out}$ guides the mean embedding of instances to repel each other. We denote $\mathcal{L}_{reg}$ as a regularization term that bounds the embedding values.
The three loss terms are denoted as:
\begin{equation}
 \begin{array}{l}
  {\mathcal{L}_{in} = \frac{1}{I}\sum\limits_{i=1}^I\frac{1}{N_i}\sum\limits_{j=1}^{N_i}{[{||{\tau}_i-f_j||}_1-{\zeta}_v]}^2_+}
 \end{array}
 \label{eq2}
\end{equation}
\begin{equation}
 \begin{array}{l}
  {\mathcal{L}_{out} = \frac{1}{I(I-1)}
  \mathop{\sum\limits_{i_a=1}^{I}\sum\limits_{i_b=1}^{I}}\limits_{i_a\not=i_b}
  {[2{\zeta}_d-{||{\tau}_{i_a}-{\tau}_{i_b}||}_1]}^2_+}
 \end{array}
 \label{eq3}
\end{equation}
\begin{equation}
 \begin{array}{l}
  {\mathcal{L}_{reg} = \frac{1}{I}\sum\limits_{i=1}^{I}{||{\tau_i}||}_1}
 \end{array}
 \label{eq4}
\end{equation}
where $I$ represents the number of ground-truth instances; $N_i$ is the number of points in instance $i$; $\tau_i$ denotes the mean embedding of instance $i$; $f_j$ is an embedding of a point; $\zeta_v$ and $\zeta_d$ indicate margins for the variance and distance loss respectively; $i_a$ and $i_b$ represent different instances; $[x]_+ =$$\rm max$$(0,x)$ is the hinge function; and the $l_1$ distance is represented by ${||\cdot||}_1$.

For inference, we use mean-shift clustering \cite{comaniciu2002mean} on the instance embeddings to obtain the final instance labels following \cite{wang2019associatively} .
The mode of the semantic labels for the points within the same instance is assigned as the predicted semantic class.
\subsection{Multi-scale Semantic Association Module}
\label{subsec:msa}
In 3D semantic segmentation, for a point of an object, the categories of surrounding points are usually related to the category of the point itself, \emph{i.e.}, the spatial distribution of the semantic information has a strong association as the ensample in Sec \ref{sec1}, which can be further exploited. Thus, based on the semantic context information, we propose our Multi-scale Semantic Association (\textit{MSA}) Module, which can be seen in Figure \ref{fig2}.\\
As shown in Figure \ref{fig2}, on the one hand, we create multi-scale semantic association maps by statistics with ball query upon all the training 3D scenes, $M_s \in \mathbb{R}^{C \times C}$ means the map in scale $s$, $C$ is the number of class. On the other hand, based on the decoded semantic output feature $P_{sem}$, we can also generate the probabilities $P_{corr}^{s}$ of the categories from surrounding points with ball query in scale $s$. Then for each point $i$ in $P_{corr}^{s}$, we calculate the distance between $P_{corr}^{s}(i)$ and each line in $M_s$, and transfer the result as a probability vector for this point, where the larger a bit is, the higher the probability for this point belonging to corresponding category is. Note that the \textit{MSA} module will generate multiple probability vectors because of multiple scales and these probabilities are only come from surrounding points. At last, the original predicted probability vector is added by the multiple probability vectors to get the final prediction. The formula is described as equation \eqref{eq6}-\eqref{eq9}
\begin{equation}
    \begin{array}{l}
     {O_{sem} = o(argmax(P_{sem}))}
    \end{array}
    \label{eq6}
\end{equation}
\begin{equation}
    \begin{array}{l}
     {\mathcal{P}_{corr}^{s}(i) = \frac{1}{|B_{i}^s|}
     \mathop{\sum\limits_{\substack{j=1\\j \in B_{i}^s}}^{|B_{i}^s|}}{O_{sem}(j)}}
    \end{array}
    \label{eq7}
\end{equation}
\begin{equation}
    \begin{array}{l}
     {\mathcal{P}_{surr}^{s}(i) = {\phi}(1-{\sigma}(||\mathcal{P}_{corr}^{s}(i)-M_s||^2))}
    \end{array}
    \label{eq8}
\end{equation}
\begin{equation}
    \begin{array}{l}
     {\mathcal{P}_{out} = \mathcal{P}_{sem}+ {{\alpha}_1}\mathcal{P}_{surr}^{s1} + {{\alpha}_2}\mathcal{P}_{surr}^{s2} + {{\alpha}_3}\mathcal{P}_{surr}^{s3}}
    \end{array}
    \label{eq9}
\end{equation}
where $o$ means one hot operation, $|B_{i}^s|$ means the number of points in the ball query of point $i$ in scale $s$, $M_s$ means the semantic association map in scale $s$, ${\sigma}$ means normalization and ${\phi}$ means softmax operation. Note that 
$\mathcal{P}_{corr}^{s}(i) \in \mathbb{R}^{1 \times C}$ and $M_s \in \mathbb{R}^{C \times C}$ can be operated with broadcast mechanism, and $||\cdot||^2$ is operated in axis 1. The final probability output is the sum of $P_{sem}$ and ${\mathcal{P}_{corr}}$ in different scales with different coefficients. In our experiment, we set $s_1,s_2,s_3$ equal to radius 0.2, 0.3, 0.5 m and ${\alpha}_{1},{\alpha}_{2},{\alpha}_{3}$ equal to 0.5, 0.3, 0.2 respectively.

\subsection{Salient Point Clustering Optimization Module}
\label{subsec:spco}
As explained in the baseline framework, for instance segmentation, the goal of $\mathcal{L}_{in}$ is to pull the embeddings toward the mean embedding of the points from the same object, while $\mathcal{L}_{out}$ guides the mean embedding of instances to repel each other in the training process. In the inference time, mean shift clustering algorithm is utilized to distinguish points of different objects. However, the coupling between the convergence orientations in training and the clustering orientation in inference is not taken into consideration. In addition, the points from the same object have different difficulties in instance segmentation as the ensample in Sec \ref{sec1}. Thus, in this paper, we propose a Salient Point Clustering Optimization (\textit{SPCO}) module, which takes mean shift clustering algorithm into the training process and saliently focuses on the points that are harder to be distinguished in the clustering process.
More specifically, as shown in Figure \ref{fig3}, mean shift clustering algorithm is operated in training process to simulate the clustering procedure in inference. Then for the points clustered in one instance while are not belonging to this instance according to the ground truth, we generate an additional loss ${\mathcal{L}_{cluster}}$ to repel these embeddings away from the mean embedding of this instance. The loss ${\mathcal{L}_{cluster}}$ is formulated in equation \eqref{eq13}, note that the ID of the clustered instance is decided by the mode of ID in the ground truth, and to converge on a reliable model, we add $\mathcal{L}_{cluster}$ into the training process from 10 epochs.
\begin{equation}
    \begin{array}{l}
     {\mathcal{L}_{cluster} = \frac{1}{N_c}
     \mathop{\sum\limits_{i=1}^{N_c}\frac{1}{|W^i|}\sum\limits_{\substack{j=1\\j \in W^i}}^{|W^i|}}
     {[2{\zeta}_d-{||{E}_{j}^i-{\overline{E}}^{i}||}_1]}^2_+}
    \end{array}
    \label{eq13}
\end{equation}
\begin{equation}
    \begin{array}{l}
     {\mathcal{L}_{ins}} = \mathcal{L}_{base} + \mathcal{L}_{cluster}
    \end{array}
    \label{eq10}
\end{equation}
where $N_c$ means the number of instances in clustering, $|W^i|$ means the number of wrong clustered points in instance $i$, ${E}_{j}^i$ means  the embedding of $j$th  wrong clustered point in instance $i$ and $\overline{E}^{i}$ means the mean embedding of the correct clustered points in instance $i$.
Equipped with our \textit{SPCO} module, the network can simulate the clustering procedure in inference more realistically, and pay more attention to the points that are easy to be erroneously clustered, which is significant for improving the performance of instance segmentation.
\subsection{Water Filling Sampling algorithm}
\label{subsec:wfs}
In indoor scenes, there exists some inherent structures. For example, the space is always surrounded by walls and floors. When we sample point clouds from indoor scenes, points of certain categories will occupy the main proportion, which will cause serious imbalance problem between these main categories and other normal categories, especially for tiny objects. In previous works of points segmentation task, this problem is rarely discussed. Therefore, in this paper, a Water Filling Sampling algorithm is proposed to solve the imbalance problem in indoor scenes, which is adaptive to different category distribution. Specifically, for a point cloud of a scene, we first cut it into blocks along $X$-$Y$ plane and store corresponding semantic and instance labels for each point in the blocks. In addition, we define an accumulative vector $VB \in \mathbb{R}^{1 \times C}$ to store the block number for each category, and generate a list $SemB[i]$ to indicate which block contains points of category $i$. If the number of points in a block that belongs to category $i$ is larger than a thresh $t$, the block index will be contained in $SemB[i]$ and $VB[i]$ will be added by 1. When we accomplished the cutting step, we can get the probabilities of block number for each category from $VB$. 
To keep the balance among the categories, we need to sample the same size of blocks from all the blocks with different probabilities. If the original probability of a category is high in the row data, the sample probability should be correspondingly low. To achieve this goal, we gradually add a small probability value $\delta$ to the category with the minimum sum of original probability and current sampling probability, until the sum of the total sampling probability values up to 1. The process is likely to fill water to the canyon consisting of original probabilities of all the categories, the details of the algorithm can be formulated as Algorithm \ref{algorithm1}. As for part segmentation datasets, such as ShapeNet, the algorithm becomes more concise because we can obtain $SemB$ and $VB$ for each object directly and skip the cutting step. Note that because of the characteristic of part segmentation dataset, we perform $WFS$ algorithm on super categories.
\begin{algorithm}
  \caption{Details of Water Filling Sampling algorithm ($WFS$)}
  \textbf{Input}: Training point clouds of all the scenes $S$ with corresponding semantic-instance labels, and a series of parameters, including threshold $t$, number of points for each block $Np$ and number of categories $Nc$.  \\
  \textbf{Output}: All the balanced blocks $Blk$ with corresponding semantic labels $SLab$ and instance labels $ILab$.\\
  \textbf{initialization}: $SemB=[[]]*Nc$,$VB=[0]*Nc$,$SP=[0]*Nc$,
  $SPB=[]$,$B=[]$,$\Omega=0$,$\delta=0.0001$
  \begin{algorithmic}[1]
    \For{ $S_i$ in all the scenes $S$}
    \State Cut $S^i$ into blocks $B^i$ along $X$-$Y$ plane.
    \State In each block, random sample $Np$ points with labels.
    \For{ $B_j^i$ in all the blocks $B^i$}
    \State $SLab_j^i\leftarrow$ Separate out corresponding labels.
    \For{ $c$ in range $[0,Nc-1]$}
    \State $pc$ = $sum(ISLab_j^i == c)$
    \If { $pc > t$}
    \State $SemB[c] \leftarrow SemB[c]$ extended with $[(i,j)]$
    \State $VB[c]$ $+= 1$
    \EndIf
    \EndFor
    \State $B \leftarrow$ = $B$ extended with $B_j^i$
    \EndFor
    \EndFor
    \State Get the original probability $OP$ = $VB / sum(VB)$
    \While{$\Omega<1$}
    \State $idx = argmin(OP)$
    \State $OP[idx]$ $+=$ $\delta$
    \State $SP[idx]$ $+=$ $\delta$
    \State $\Omega += \delta$
    \EndWhile
    \For{ $c$ in range $[0,Nc-1]$}
    \State $Sc = SP[c]* length(B) $
    \State $Bc \leftarrow$ Random sample $Sc$ block indicates in $SemB[c]$
    \State $SPB \leftarrow$ $SPB$ extended with $Bc$ 
    \EndFor
    \State $B \leftarrow$ $B$ extended with $B[SPB]$ 
    \State Separate $B$ into $Blk$, $SLab$ and $ILab$
    \State Return $Blk$, $SLab$, $ILab$
  \end{algorithmic}
  \label{algorithm1}
\end{algorithm}
\section{Experiments}\label{sec4}
In this part, we will compare our method with other SOTA methods in 3D point clouds semantic and instance segmentation tasks to demonstrate that our method is effective and robust on different kind of datasets, including large scale indoor 3D dataset and part segmentation 3D dataset.
\subsection{Datasets and Details}
\noindent \textbf{Datasets.} Followed as \cite{wang2019associatively}, we conduct the experiments on two benchmark datasets: Stanford 3D Indoor Semantics Dataset (S3DIS) \cite{armeni20163d} and ShapeNet part segmentation Dataset \cite{yi2016scalable}. The specific introduction of these datasets is as follows:
\begin{itemize}
\item S3DIS is a real 3D point cloud dataset generated by Matterport Scanners for indoor spaces, which contains 6 areas and 272 rooms. Each point contains 9 dimensions for the input feature including $XYZ$, $RGB$ and normalized coordinates. For each point, an instance ID and a semantic category ID with 13 classes are annotated. Following \cite{qi2017pointnet}, we split the rooms into 1 m $\times$ 1 m overlapped blocks with stride 0.5 m along the $X$-$Y$ plane and sample 4096 points from each block.
\item ShapeNet dataset is a synthetic scene mesh for part segmentation, which consists of 16881 shape models from 16 categories. Each object is annotated with 2 to 5 parts from 50 different sub-categories. We utilize the instance annotations generated by \cite{wang2018sgpn} as the ground-truth labels and we sample 2048 points for each shape during training followed as \cite{qi2017pointnet}. We split the dataset into training and validation followed \cite{wang2019associatively} and 3-dimensional vector including $XYZ$ is fed into our network as input.
\end{itemize}
\begin{figure*}[h]
  \begin{center}
   \centerline{\includegraphics[width=18cm]{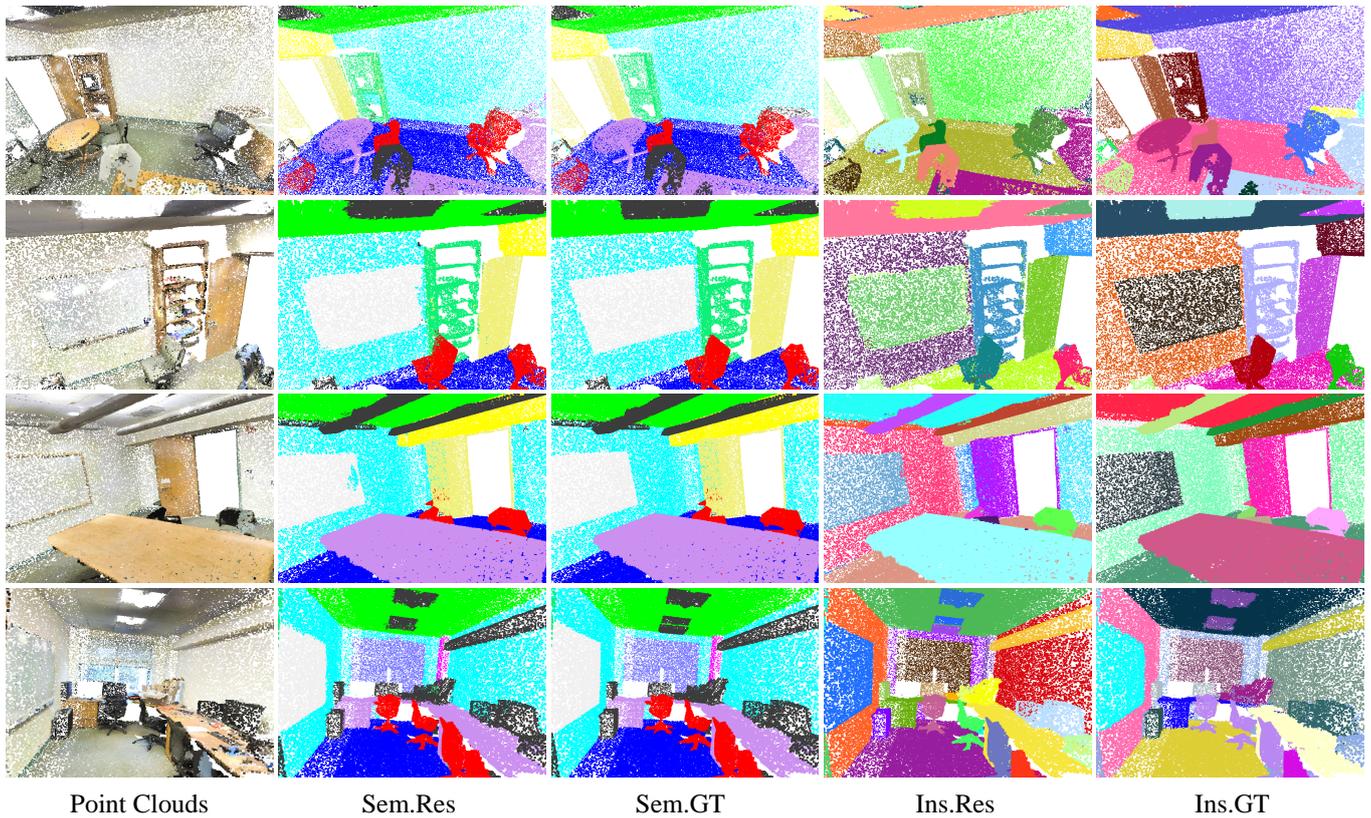}}
  \end{center}
  \caption{Qualitative results of our method on the S3DIS dataset. For semantic results, each color refers to a particular category and for instance results, different colors represent different objects. }
  \label{fig4}
\end{figure*}
\begin{table}[htb]
  \caption{Semantic (green) and instance (red) segmentation results on S3DIS.}
  \centering
  \LARGE
  \resizebox{86mm}{18mm}{
   \begin{tabular}{c|c|cccc|ccc}
  \hline
   Dataset\cellcolor{mygray} &Method \cellcolor{mygray} & mCov \cellcolor{mypink} & mWCov \cellcolor{mypink} & mPrec \cellcolor{mypink} & mRec \cellcolor{mypink} & mAcc \cellcolor{mygreen} & mIou \cellcolor{mygreen} & oAcc \cellcolor{mygreen} \\
    \hline
    \multirow{4}{*}{Area5} &SGPN \cite{wang2018sgpn} & 32.7 & 35.5 & 36.0 & 28.7  &\multicolumn{3}{c}{------------------------}   \\
    &JSIS3D \cite{pham2019jsis3d} & 32.6 & 35.6 & 39.7  & 29.1 & 59.2 & 51.8  & 86.9  \\
    &3D-BoNet \cite{yang2019learning} & 41.5 & 44.6 & 57.6  & 40.2 & 59.2 & 51.8 & 86.9\\
    &ASIS \cite{wang2019associatively}   & 44.6  & 47.8 & 55.3  & 42.4 & 60.9 & 53.4  & 86.9  \\
    &OURS              & \textbf{49.0}           & \textbf{51.9}            & \textbf{59.5}            & \textbf{45.9}           & \textbf{63.5}            & \textbf{55.5}            & \textbf{87.5}            \\
  \hline
  \hline
  \multirow{4}{*}{6-Fold CV} &SGPN \cite{wang2018sgpn} &37.9 &40.8 &38.2 &31.2 &\multicolumn{3}{c}{------------------------}   \\
    &JSIS3D \cite{pham2019jsis3d}  &37.3 &41.0 &49.5  &33.4 &59.8 &48.5  &79.9  \\
    &3D-BoNet \cite{yang2019learning}&48.4 &52.4 &\textbf{65.6}  &47.6 &69.3 &59.4  &86.3  \\
    &ASIS \cite{wang2019associatively}&51.2  &55.1 &63.6  &47.5 &70.1 &59.3  &86.2  \\
    &OURS              & \textbf{54.5}           & \textbf{58.3}            & 64.2            & \textbf{50.8}           & \textbf{72.8}            & \textbf{61.1}            & \textbf{87.0}            \\
    \hline
  \end{tabular}}
  \label{tab1}
\end{table}
\begin{table}[htb]
  \caption{Ablation study on the S3DIS dataset in Area5.}
  \tiny
  \centering
  \resizebox{86mm}{13mm}{
   \begin{tabular}{ccc|cc|cc}
  \hline
    SPCO\cellcolor{mygray}&MSA\cellcolor{mygray} &WFS\cellcolor{mygray} & mWCov \cellcolor{mypink} & mPrec \cellcolor{mypink} & mAcc \cellcolor{mygreen} & mIou \cellcolor{mygreen}\\
    \hline
    $\times$&$\times$&$\times$       &47.1 &51.9 &59.7 &52.0      \\
    \checkmark&$\times$&$\times$     &50.3 &56.0 &61.6 &53.6      \\
    $\times$&\checkmark&$\times$       &47.1 &51.9 &61.4 &53.2      \\
    $\times$&$\times$&\checkmark       &49.8 &55.3 &61.2 &53.3      \\
    \checkmark&\checkmark&$\times$       &50.3 &56.0 &62.7 &54.5      \\
    \checkmark&\checkmark&\checkmark       &\textbf{51.9} &\textbf{59.5} &\textbf{63.5} &\textbf{55.5} \\
  \hline
  \end{tabular}}
  \label{tab2}
\end{table}
\begin{table*}[htb]
  \caption{Per class results on the S3DIS dataset.}
  \tiny
  \centering
  \resizebox{180mm}{12mm}{
   \begin{tabular}{c|c|c|ccccccccccccc}
    \hline
    Metrics \cellcolor{mygray} &Method &mean &ceiling &floor &wall &beam &column &window &door &table &chair &sofa &bookcase &board &clutter\\
    \hline
    \cellcolor{mypink} & BASE         & 47.1             &\textbf{89.7}          & 88.7          & 68.3          & 0.0           & 3.4          &\textbf{60.9}           & 5.0       & 51.8          & 67.6              & 23.9          & 53.6 & 50.3   & \textbf{49.5}                 \\
    \cellcolor{mypink}  Wcov  & ASIS$^{*}$ \cite{wang2019associatively}   &47.6 &89.0 &\textbf{89.2} &72.4 &0.0 &8.8 &58.1 &4.7 &52.4 &\textbf{76.6}  &46.3  &50.1 &64.4 &45.5          \\
    \cellcolor{mypink}         & OURS       &\textbf{51.9} &89.0 &87.3 &\textbf{73.1} &\textbf{0.0} &\textbf{9.1} &60.1 &\textbf{13.3} &\textbf{54.3} &69.8 &\textbf{48.7} &\textbf{55.0} &\textbf{68.1} &46.6 \\
    \hline
  \cellcolor{mygreen}        & BASE  & 52.0          & \textbf{92.8}         & 97.8          & 74.8          & 0.0           & 7.9          & \textbf{51.9}          & 16.1          & 72.3                     & 77.9               & 35.4 & 56.1          & 42.5          & 50.8          \\
    Sem IoU \cellcolor{mygreen}   & ASIS$^{*}$ \cite{wang2019associatively}      & 53.4   & 92.4 &\textbf{98.4} &76.7 &0.0 &\textbf{15.6} &49.5 &21.4 &72.3 &78.7 &38.0 &55.9 &45.8  &49.7\\
    \cellcolor{mygreen}        & OURS       &\textbf{55.5} &92.5 &97.7 &\textbf{77.2} &\textbf{0.0} &11.7 &50.8 &\textbf{29.0} &\textbf{74.2} &\textbf{80.3} &\textbf{41.3} &\textbf{60.0} &\textbf{56.6} &\textbf{50.8}\\
   \hline
   \end{tabular}}
  \label{tab3}
\end{table*}
\begin{table}[h]
  \caption{Comparisons of computation time, GPU memory and performance.}
  \tiny
  \centering
  \resizebox{84mm}{10mm}{
   \begin{tabular}{c|cc|cc|c}
   \hline
    \multirow{2}{*}{\diagbox{Method}{Metrics}}              & \multicolumn{2}{c|}{Train}       & \multicolumn{2}{c|}{Test} &  \multirow{2}{*}{mPrec}  \\
    \cmidrule{2-5}
    & time (m)      & memory (MB)     & time (m)      & memory (MB)   \\
    \hline
    SGPN \cite{wang2018sgpn}& \textbf{59.3}          & 7549          & 209.5         & 420          & 36.0          \\
    ASIS \cite{wang2019associatively} & 64.7          & 4275          & 54.2          & 1235         & 55.3          \\
    OURS & 75.0 & \textbf{1203} & \textbf{40.4} & \textbf{373} & \textbf{59.5} \\
 \hline
   \end{tabular}}
  \label{tab4}
\end{table}
\noindent \textbf{Details.}
For instance segmentation, we trained SASO with $\lambda = 0.001$. We use five output embeddings following \cite{wang2019associatively} and set $\alpha$ to 0.01. We select the Adam optimizer to optimize the network on a single GPU (Tesla P100) and set the momentum to 0.9 for the training process. During the inference process, we set the bandwidth to 0.6 for mean-shift clustering and apply the BlockMerging algorithm \cite{wang2018sgpn} to merge instances from different blocks.\\ \hspace*{\fill} \\
\noindent \textbf{Evaluation.} Following \cite{wang2019associatively}, we evaluate the experimental results in the following metrics. For semantic segmentation, we calculate the overall accuracy ($oAcc$), mean accuracy ($mAcc$) and mean $IoU$ ($mIoU$) across all the semantic classes along with the detailed scores of the per-class $IoU$. To evaluate the performance of instance segmentation, we use the coverage ($Cov$) and weighted coverage ($WCov$) \cite{ren2017end, liu2017sgn, zhuo2017indoor}.
$Cov$ is the average instance-wise IoU of the prediction matched with ground truth, and $WCov$ is the $Cov$ score after being weighted by the size of ground truth.
For the predicted regions $P$ and the ground-truth regions $G$, $Cov$ and $WCov$ are defined as:
\begin{equation}
 \begin{array}{l}
  {{\rm Cov}(\mathcal{G},\mathcal{P})=\sum\limits_{i=1}^{|\mathcal{G}|}\frac{1}{|\mathcal{G}|}\max\limits_{j}{\rm IoU}(r_i^{\mathcal{G}}, r_j^{\mathcal{P}})}
 \end{array}
 \label{eq10}
\end{equation}
\begin{equation}
 \begin{array}{l}
  {{\rm WCov}(\mathcal{G},\mathcal{P})=\sum\limits_{i=1}^{|\mathcal{G}|}w_i\max\limits_{j}{\rm IoU}(r_i^{\mathcal{G}}, r_j^{\mathcal{P}})}
 \end{array}
 \label{eq11}
\end{equation}
\begin{equation}
 \begin{array}{l}
  {{w_i} = \frac{|r^{\mathcal{G}}_i|}{{\sum_k}|r^{\mathcal{G}}_k|}}
 \end{array}
 \label{eq12}
\end{equation}
where $|r_i^{\mathcal{G}}|$ is the number of points in ground-truth region $i$.
We also measure the classical metrics of mean precision ($mPrec$) and mean recall ($mRec$) with an $IoU$ threshold of 0.5. \\
\begin{figure}[h]
  \begin{center}
   \centerline{\includegraphics[width=8.5cm]{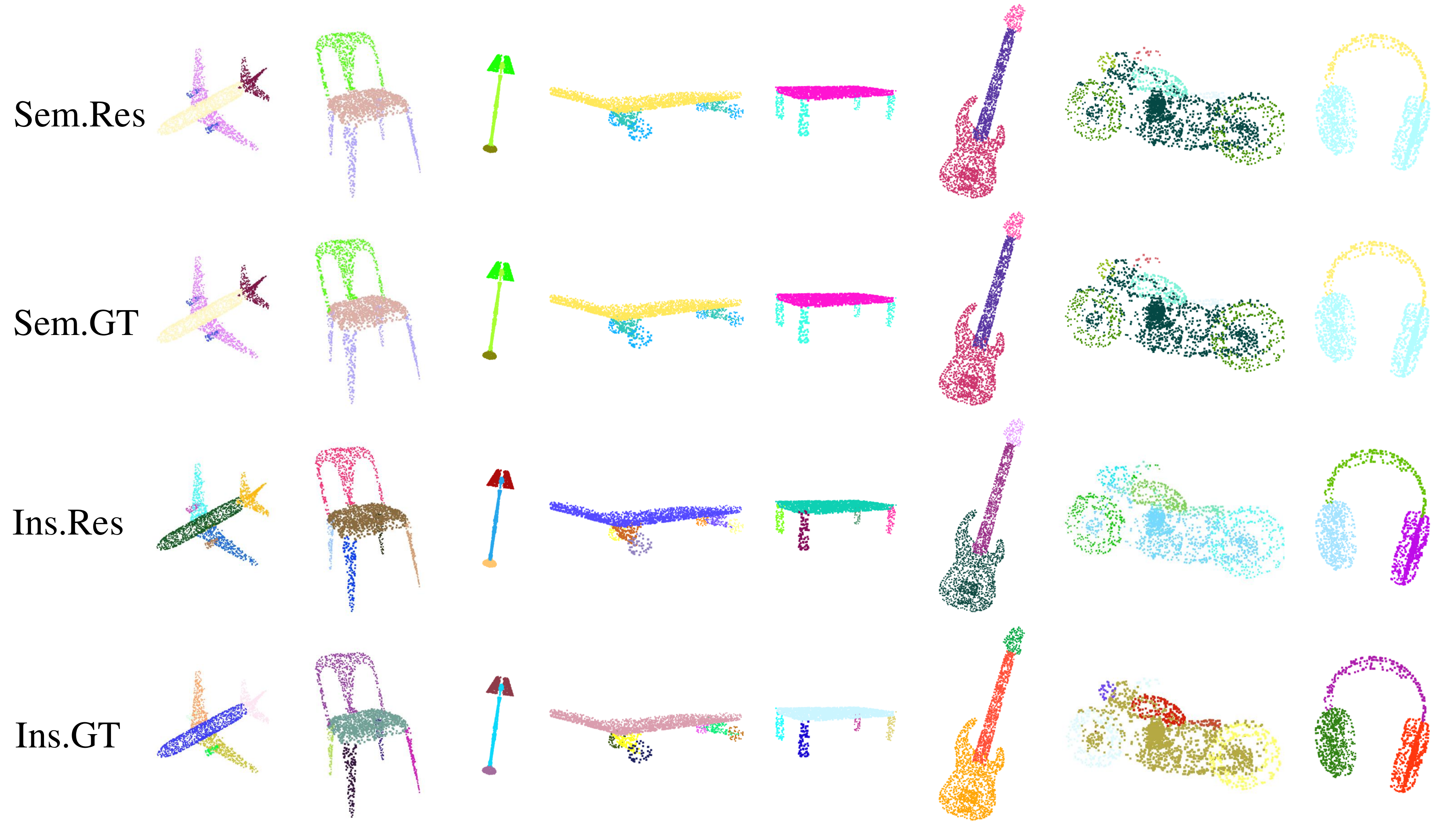}}
  \end{center}
  \caption{Qualitative results for semantic and instance segmentation on ShapeNet dataset.}
  \label{fig6}
\end{figure}
\subsection{S3DIS Evaluation}
We conduct the experiments on the S3DIS dataset with the backbone networks PointNet++. We train the network for 50 epochs with a batch size of 12, the initial learning rate is set to 0.001 and divided by 2 every 300 k iterations.\\ \hspace*{\fill} \\
\textbf{Quantitative Results.} For classical Area5 validation scenes, the quantitative results of SASO in instance and semantic segmentation tasks are shown in Table \ref{tab1}. As we can see, SASO achieves 51.9 $mWCov$ and 59.5 $mPrec$, which dramatically outperforms the state-of-the-art method 3D-BoNet \cite{yang2019learning} by 7.3 in $mWCov$ and 1.9 in $mPrec$. As for semantic segmentation, our method significantly improves the $mAcc$ and $mIoU$ by 2.6 and 2.1 respectively, compared with advanced ASIS \cite{wang2019associatively}.
For a more comprehensive comparison, we evaluate our method with 6 fold cross validation on S3DIS dataset. As shown in the table,
our method achieves 58.3 $mWCov$ and 72.8 $mAcc$, which significantly outperforms the state-of-the-art methods by a large margin.
\begin{figure}[t]
  \begin{center}
   \includegraphics[width=8.5cm]{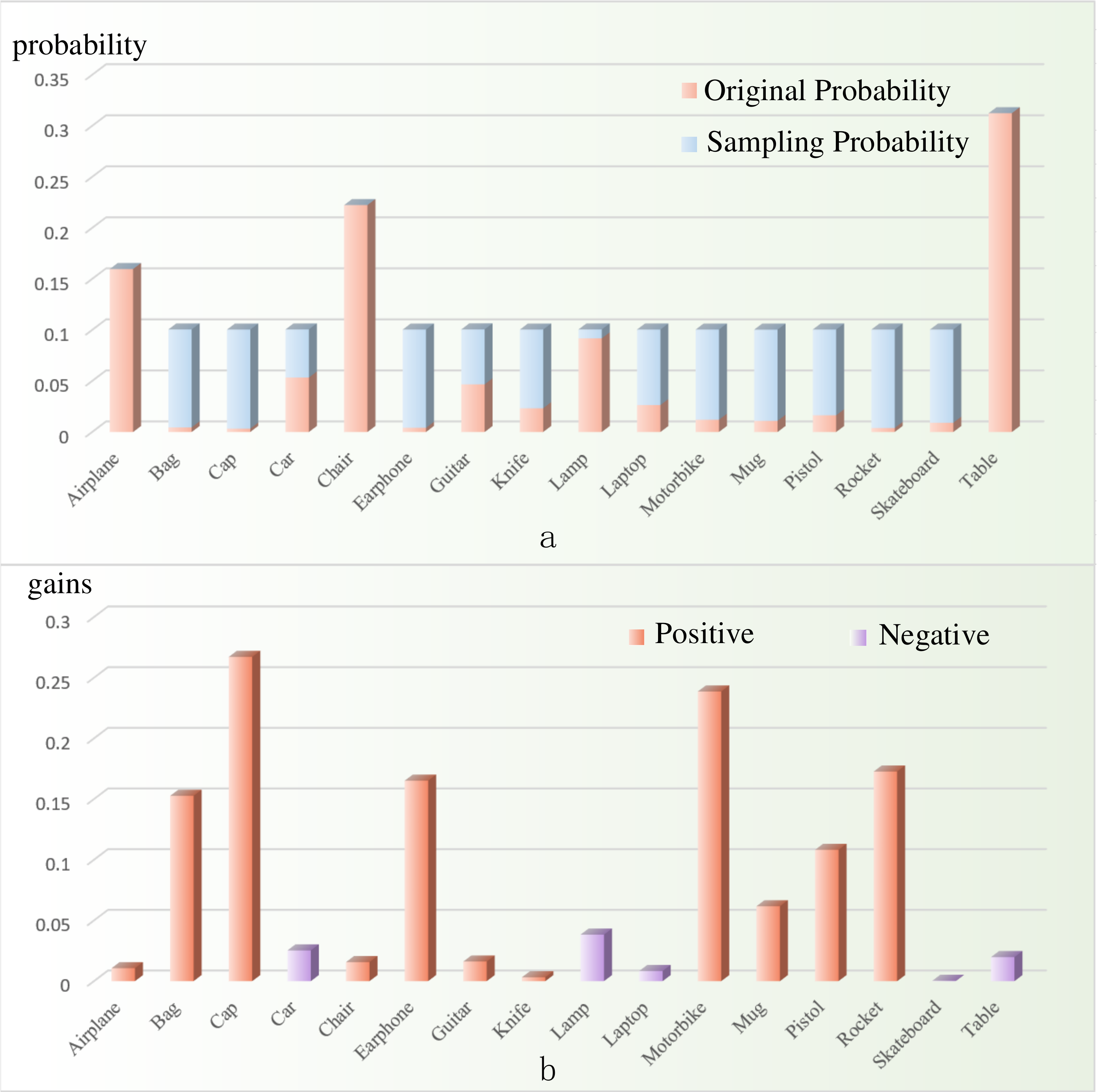}
  \end{center}
  \caption{The sampling probability and the corresponding improvement for different categories upon ShapeNet dataset.\\
  (a). The orange color means the original frequency for different categories in the training dataset, the blue color represents the sampling probabilities for different categories.\\
  (b). The orange color means positive boost while the purple color represents negative influence. Note that for an intuitive visualization, the value are multiplied by 5. }
  \label{fig5}
  \end{figure}
The stable improvement in both semantic and instance segmentation demonstrates the effectiveness of our method.
For a more detailed comparison with our baseline framework and ASIS \cite{wang2019associatively}, Table \ref{tab3} shows the results for specific categories in both instance and semantic segmentation based on Area5 scene in S3DIS. Note that for a fair comparison, we reproduce the result of ASIS \cite{wang2019associatively} with PointNet++ backbone using the author's code to get the per class results.\\ \hspace*{\fill} \\
\textbf{Qualitative Results.} To intuitively present our results, we visualize the predict results and annotations on point clouds, as shown in Figure \ref{fig4}. For instance segmentation, different colors represent different instances. For semantic segmentation, each color refers to a particular category. It is obvious that our method has a great performance, especially at the boundaries of different objects.\\ \hspace*{\fill} \\
\textbf{Ablation Study.} The ablation study results are shown in Table \ref{tab2}. Equipped with different modules of our method upon the baseline framework, we can find that with our $SPCO$ module, we obtain 3.2 gains in $mWCov$ and 4.1 gains in $mPrec$. It is interesting that the semantic segmentation results are also improved with this module, we think this is because the semantic and instance segmentation tasks share the shallow features, the improvement in the instance segmentation branch can be beneficial to semantic segmentation branch. When we add $MSA$ module to the baseline, we can find that the semantic segmentation results are improved with 1.7 in $mAcc$ and 1.2 in $mIoU$. With the \textit{WFS} algorithm added to the baseline framework, we obtain 3.4 gains in $mPrec$ and 1.3 gains in $mIoU$, which means the balance among different categories is critical to both two tasks. Finally, compared with the baseline framework, our full method has a dramatic improvement in both two tasks, including 7.6 $mPrec$ gains in instance segmentation task and 3.5 $mIoU$ gains in the semantic segmentation task.\\ \hspace*{\fill} \\
\textbf{Consumption of memory and time.} Table \ref{tab4} shows a comparison of the memory cost and computation time. For a fair comparison, we conducted the experiments in the same environment, including the same GPU (GTX 1080), batch size (4) and data (Area5 including 68 rooms).
Note that all the time units are minutes, and all the memory units are MB. In the training process, the result is the time and memory required for one epoch. As we can see, our method needs relatively more time for training because we introduce clustering into training process, while costs little memory because of the brief but efficient architecture. In the inference process, the results show the resource consumption for Area5. Our approach takes only 373 MB and needs 40.4 minutes while acquires better performance, which is significantly faster and more efficient than the state-of-the-art methods.
\begin{table}[htb]
  \caption{Semantic segmentation results on ShapeNet datasets.}
  \tiny
  \centering
  \resizebox{32mm}{16mm}{
  \begin{tabular}{c|c}
 \hline
   \textbf{Method}        & mIoU \cellcolor{mygreen} \\
   \hline
   PointNet++ \cite{qi2017pointnet++} & 84.3                     \\
   ASIS \cite{wang2019associatively}                             & 85.0                     \\
   SGPN\cite{wang2018sgpn}                             & 85.8                     \\
   SpiderCNN \cite{xu2018spidercnn} &85.3\\
   SSCN \cite{graham20183d} &86.0 \\
   PointConv \cite{wu2019pointconv} &85.7 \\
   BASE  &83.5 \\
   OURS                            & \textbf{86.4}           \\
 \hline
  \end{tabular}}
  \label{tab5}
\end{table}
\subsection{ShapeNet Evaluation}
We also validate our method on part segmentation dataset ShapeNet, the semantic annotations are publicly available while the instance segmentation annotations are the generated results as \cite{wang2018sgpn}. Because of the deficiency of ground truth for instance annotations, we only provide the qualitative results for instance segmentation in Figure \ref{fig6} as \cite{wang2019associatively}. Four lines from top to bottom in Figure \ref{fig6} mean semantic segmentation results, semantic annotations, instance segmentation results and instance annotations respectively.
As we can see, different parts in the same object are well grouped into individual instances, especially the boundaries of different parts.
The semantic segmentation results are exhibited in Table \ref{tab5}. Our approach obviously boosts the result upon baseline framework by 2.9 $mIoU$ and outperforms the state-of-the-art method ASIS \cite{wang2019associatively}, PointConv \cite{wu2019pointconv} and SSCN \cite{graham20183d}.
These results reveal that our proposed method also has the capability to improve the part segmentation performance.\\
To prove the effectiveness of our \textit{WFS} algorithm intuitively, we show the sampling probability and the corresponding improvement for different categories, as depicted in Figure \ref{fig5}. In the upper graph (a), the orange color means the original frequency of different categories in the training dataset, the blue color represents the sampling probabilities for different categories. We can find that the distribution of different categories is more balanced with our \textit{WFS} algorithm. The second graph (b) shows the improvement for different categories, the orange color means positive boost while the purple color represents negative influence. For the categories with low frequency existing in the raw data, the corresponding improvements are obvious, while for the categories with high frequency, the results are rarely influenced. It demonstrates that our \textit{WFS} algorithm is effective and critical for alleviating the imbalance problem.
\section{Conclusion} 
In this paper, we propose a novel framework which jointly performs semantic and instance segmentation. For the instance segmentation task, a module named \textit{SPCO} is proposed to introduce clustering into the training process and saliently focus on the points that are harder to be distinguished in the clustering process. For the semantic segmentation branch, we introduce \textit{MSA} module based on the statistic knowledge to exploit the potential association of spatial semantic distribution. In addition, we propose a Water Filling Sampling algorithm to address the imbalance problem of category distribution. Qualitative and quantitative experiment results on challenging benchmark datasets demonstrate the effectiveness and robustness of our method.

\section*{Acknowledgment}
* This project was supported by National Natural Science Foundation of China (No.61806189) and Shanghai Municipal Science and Technology Major Project (Grant No. 2018SHZDZX01, ZHANGJIANG LAB).

\bibliographystyle{iet}
\bibliography{sample}

\end{document}